\documentclass[conference]{IEEEtran}
\IEEEoverridecommandlockouts
\usepackage{cite}
\usepackage{amsmath,amssymb,amsfonts}
\usepackage{algorithmic}
\usepackage{graphicx}
\usepackage{textcomp}
\usepackage{xcolor}
\usepackage{comment}
\usepackage{multirow} 
\usepackage{footnote}
\usepackage{url}
\def\BibTeX{{\rm B\kern-.05em{\sc i\kern-.025em b}\kern-.08em
    T\kern-.1667em\lower.7ex\hbox{E}\kern-.125emX}}
\begin{document}

\title{Extraction of Medication and Temporal Relation from Clinical Text using Neural Language Models 
}

\author{\IEEEauthorblockN{Hangyu Tu}
\IEEEauthorblockA{\textit{Department of Computer Science} \\
\textit{University of Manchester}\\
Manchester, UK \\
hangyu.tu@student.manchester.ac.uk}
\and
\IEEEauthorblockN{Lifeng Han}
\IEEEauthorblockA{\textit{Department of Computer Science} \\
\textit{University of Manchester}\\
Manchester, UK \\
lifeng.han@manchester.ac.uk}
\and
\IEEEauthorblockN{Goran Nenadic}
\IEEEauthorblockA{\textit{Department of Computer Science} \\
\textit{University of Manchester}\\
Manchester, UK \\
g.nenadic@manchester.ac.uk}
}

\maketitle

\begin{abstract}
Clinical texts, represented in electronic medical records (EMRs), contain rich medical information and are essential for disease prediction, personalised information recommendation, clinical decision support, and medication pattern mining and measurement. 
Relation extractions between medication mentions and temporal information can further help clinicians better understand the patients' treatment history.
To evaluate the performances of deep learning (DL) and large language models (LLMs) in medication extraction and temporal relations classification, we carry out an empirical investigation of \textbf{MedTem} project using several advanced learning structures including 
BiLSTM-CRF and CNN-BiLSTM for a clinical domain named entity recognition (NER), and BERT-CNN for temporal relation extraction (RE), in addition to the exploration of different word embedding techniques. Furthermore, we also designed a set of post-processing roles to generate structured output on medications and the temporal relation.
Our experiments show that CNN-BiLSTM slightly wins the BiLSTM-CRF model on the i2b2-2009 clinical NER task yielding 75.67, 77.83, and 78.17 for precision, recall, and F1 scores using Macro Average. BERT-CNN model also produced reasonable evaluation scores 64.48, 67.17, and 65.03 for P/R/F1 using Macro Avg on the temporal relation extraction test set from i2b2-2012 challenges. Code and Tools from MedTem will be hosted at \url{https://github.com/HECTA-UoM/MedTem}
\end{abstract}


\section{Introduction}
\label{chapt_intro}

\begin{figure*}[t]
    \centering
    \includegraphics[width=0.8\textwidth]{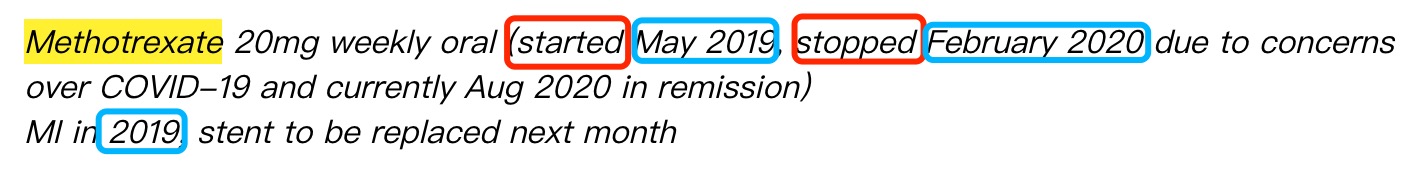}
    \caption{Original Text Example from EMRs}
    \label{fig:example_text_EMRs}
\end{figure*}

With the continuous growth of the need for medical text mining, extraction of temporal information has become a significant field of Natural Language Process (NLP) \cite{Su2004LectureScience}, e.g. when the disease was diagnosed, how long a patient should take some medicines. 
Due to its importance for many other NLP tasks, temporal information tagging and annotation from clinical texts have become an active research field (Figure \ref{fig:example_text_EMRs} for example). 
Clinical texts including electronic health records (EHRs) consist of medical history, diagnoses, medications, treatment plans, immunisation dates, allergies, radiology images, and laboratory and test results, which 
contain a lot of useful information that can provide the opportunity to support medical research \cite{han2023investigatingMMPMT,zhou2021clinicalTR}. 
These include the study of the side effects of medications (\textit{a.k.a} adverse drug effects) taken over time, helping doctors make better decisions for patients, and finding the correlation between drug use and its outcomes, etc. \cite{Fredriksen2014EffectivenessComorbidity,alrdahi2023medmine}.

 In view of the importance of temporal information in clinical free texts, in order to extract it from unstructured data and convert it to structured features consisting of temporal relations (TLINKs) between events and time expressions, researchers have developed some techniques, tools, and workflows. Generally, they can be divided into three main categories: rule-based approaches, machine learning and combined approaches.
 Most machine learning methods rely on feature-engineered supervised learning to predict temporal relations between events, which require tagged datasets that cost a lot of effort and time of experts in the medical domain.

 With the development of various deep learning (DL) models nowadays, this work aims at an empirical investigation on how they perform on this specific challenge, i.e. \textit{medication and temporal relation extraction from clinical text}.

There are two major sub-tasks: 1) Medication Entity Recognition, which involves finding and identifying named drugs mentioned in unstructured text, the first step toward automating the extraction of medications from EMRs. 2) Automatic Temporal Relation Extraction from free text EMRs. We approach these two steps by applying several advanced DL methods including the BiLSTM+CRF, CNN-BiLSTM, and BERT-base-CNN models to train on the available data sets. 
Additionally, a SparkNLP tool (DateNormalizer) \cite{kocaman2021spark} is used to extract and normalise dates from relative date-time phrases. 
 


\section{Methodological Design}
\label{chapt_methodology}

\begin{figure*}
    \centering
    \includegraphics[width=0.8\textwidth]{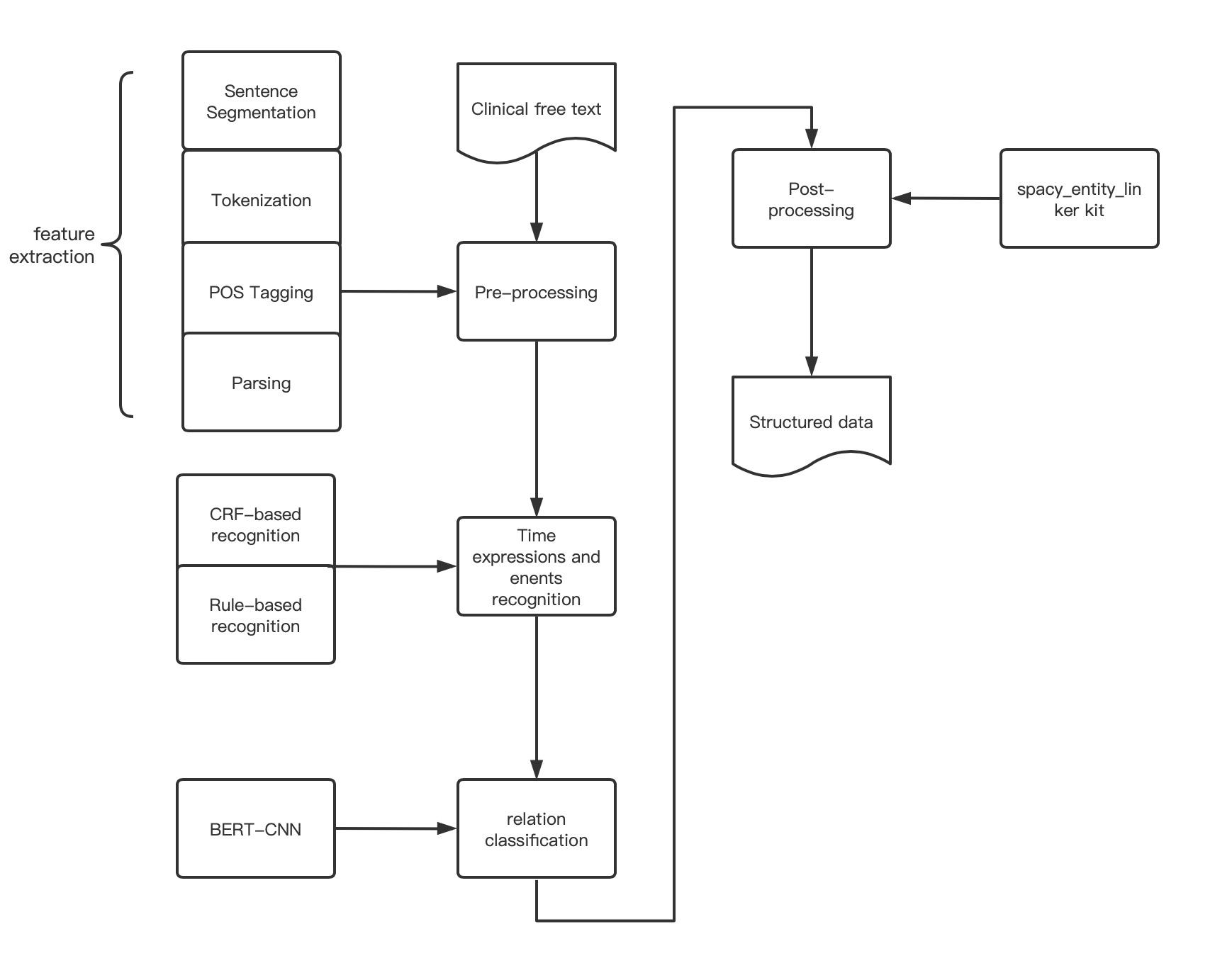}
    \caption{Methodology Illustration on Overall Tasks of MedTem}
    \label{fig:pipeline-system}
\end{figure*}

 Our designed methodology includes five main steps as shown in Figure \ref{fig:pipeline-system}, which are Data Pre-processing, Word Embedding, Modelling, Time Expression and Event Extraction, and Temporal Relations and Medications Candidates Classification.

\subsection{Pre-Processing Details}
\label{appendix_sec_preprocessing}

 Clinical texts are mixed with different formats. The target of pre-processing is trying to format different data with certain rules and enrich texts with their potential features. 
 Typically, Pre-processing follows 5 steps: Sentence Segmentation, Tokenisation, Parts of Speech (POS, which explains how a word is used in a sentence) Tagging, and Parsing. 
 For our work,  GATE \cite{GATE}, cTAKES \cite{savova2010mayo} and MedSpaCy (a new clinical text processing toolkit in Python) shows excellent power in building rule-based modules to retrieve extra features. Additionally, the library `spacy' is valuable and powerful in linking mentions to the associated verb. 

\subsection{Word Embeddings}
\label{appendix_sec_word_embedding}

 Word embedding techniques play a role in capturing the meaning, semantic relationship, and context of different words while creating word representations \cite{pham2018exploiting}. A word embedding is a technique used to generate a dense vector representation of words that include context words about their own. In addition to one-hot embedding that results in sparse matrix embedding, there are enhanced versions of simple bag-of-words models, such as word counts and frequency counters, that essentially represent sparse vectors.
\subsubsection{One-hot Encoding}

 A popular encoding technique is called ``one-hot encoding".
 In NLP, if a dictionary or 
 sequence has N fields, each field can be represented by an n-dimensional one-hot vector. 
 One-hot encoding can convert classified data into a unified digital format, which is convenient for machine learning algorithms to process and calculate since computers cannot understand non-numeric data. 
 And the fixed dimension vector is convenient for the machine learning algorithm to calculate the linear algebra.
 Data that don't relate to one another can benefit from one-hot encoding.

\subsubsection{Bag-of-words embedding}
 A bag-of-words model (BoW) is a way of extracting features from the text for use in modelling, such as with machine learning algorithms, which is a popular way of representing documents \cite{shao2018clinical}. Throughout the realm of information retrieval, the BoW model is used in information retrieval on the assumption that a text may be reduced to a collection of words without regard to its actual order within the document, its grammar, or its syntax. 
 Each word's presence in the document is autonomous and unrelated to the incidence of other terms. In other words, each word at each place in the document is individually selected without regard to the document's semantics.

\subsubsection{GloVe Embedding}

 GloVe stands for “Global Vectors” \cite{pennington2014glove}. 
 Compared to Word2vec, which only keeps local statistic information, the advantage of GloVe is that it captures both global statistics and local statistics of a corpus, and it is an unsupervised learning pre-trained model that represents words in sentences. It has the ability to extract semantic relationships.
  The main idea of the Glove method is that ``semantic relationships between words can be extracted from the co-occurrence matrix'' \cite{bullinaria2007extracting}. Given a document having a sentence of \textbf{n} words, the co-occurrence matrix \textbf{X} will be a \textbf{n}*\textbf{n} matrix.

\subsection{Deployed Learning Models}

\subsubsection{BiLSTM}

In machine learning fields, 
the problem that the model cannot get the previous moment information due to the vanishing gradient is also called the long-term dependence problem. 
In order to solve this problem, many researchers have tried to improve the Recurrent Neural Network (RNN) model, including BRNN \cite{arisoy2015bidirectional}, GRU \cite{tang2016question}, and LSTM \cite{hochreiter1997long}, etc., among which the LSTM model is the most widely used one.

 LSTM is an improved RNN structure that can effectively avoid the long-term dependence problem that the traditional RNNs cannot solve. 
 In the later work, many people adjusted and popularised the model, and the modern, widely used LSTM structure came into being. 
 LSTM network has the same chain structure as traditional RNNs, but the repeating module has a different design. The RNN model only has a single layer of repeated modules. In contrast, the LSTM network has four interacting neural network layers, which makes the LSTM model more complex and sacrifices a certain amount of computational time, but the model output becomes better. The standard LSTM model 
 mainly includes three groups of adaptive element multiplication gates, namely, the input gate, forgetting gate, and output gate.
 
In actual text processing applications, the text is characterised by sequential feature correlation, and words in the current moment are not only affected by words in the past moment but also by words in the future moment. In order to make the model contain contextual information at the same time, researchers put forward a bidirectional LSTM model (BiLSTM) \cite{zhang2015bidirectional,zhou2016attention}. BiLSTM contains two LSTM layers, which train the forward and backward sequences, respectively, and both LSTM layers are connected to the output layer. This forward-backward two-layer LSTM structure can provide historical information and future information to the output layer at the same time, which can not only retain the advantages of LSTM to solve long-term dependence but also take into account the context information and effectively deal with sequence problems.

\subsubsection{CRF model}

 Conditional random fields (CRFs) \cite{lafferty2001conditional} are a type of statistical modelling methodology frequently employed in pattern recognition and machine learning, as well as for structured prediction, which is a widely-applied modelling strategy for many NLP tasks \cite{han2013chinese,han2015chinese}. Data having an underlying graph structure can be better modelled using CRFs, a graphical model that can take advantage of structural dependencies between outputs.
In contrast to classifiers, which make label predictions for isolated data without considering their ``neighbouring'' samples, CRFs are able to account for context \cite{maldonadoHanMoreau2017detection,moreau_etal2018mwe,Han_Wu_etal2022_PLM4clinical}. In order to accomplish this goal, the predictions are modelled in the form of a graphical model, which depicts the existence of relations between the predictions. The choice of the graph to implement is context-specific. 
In NLP fields, for instance, ``linear chain'' CRFs are prevalent because each prediction depends only on its immediate neighbours. Connecting neighbouring and/or analogous regions ensures that all regions within a picture are predicted in the same way by the graph. 

\subsubsection{CNN model}

 In deep learning (DL), a classic convolutional neural network (CNN) is a class of artificial neural networks (ANN), most commonly applied to analyse visual imagery \cite{albawi2017understanding}. A CNN model mainly consists of these layers: input layer, convolution layer, relu layer, pooling layer and full connection layer.
 One of CNN's most alluring quality is its ability to make use of \textit{spatial} or \textit{temporal} correlation in data. CNN is divided into multiple learning phases, each of which is made up of a combination of convolutional layers, nonlinear processing units, and subsampling layers \cite{Goodfellow-et-al-2016}. In a CNN, each layer of the network uses a set of convolutional kernels to perform many transformations. The convolution process simplifies extracting useful features from spatially correlated data points. 
 We will use CNN to learn the character relations within word tokens for our task.

\subsubsection{BERT model}

Brief for ``Bidirectional Encoder Representations from Transformers'', BERT \cite{devlin2018bert} is founded on Transformer structure \cite{google2017attention}, a cutting-edge DL model in which all output elements are related to all input elements and the weights between them are dynamically computed based on their connections. 
BERT is unique in that it can be used with text that is read either from left to right or right to left. 
With this characteristic, BERT was originally taught for two NLP tasks: Next Sentence Prediction and Masked Language Modelling. 
  The input for BERT is comprised of Token Embeddings, Segment Embeddings, and Position Embeddings, followed by pre-training. 
  BERT enables effective dynamic feature encoding of polysemous words. In other words, the same word might produce a variety of word vector outputs depending on the language environment. Additionally, the pre-trained model's embedding layer transfers the knowledge from the corpus to make the word vectors more generic, greatly enhancing the model's accuracy.

\subsection{Time Expressions and Events Named Entity Recognition}

The goal of NER task is to mark up temporal information present in clinical text in order to enable reasoning on relevant events (medications) for each patient. 
 In this work, a commonly used format in entity tagging ``Inside–outside–beginning (IOB)'' \cite{ramshaw1999text}, is used, which is explained below:

\begin{enumerate}
    \item the B-prefix indicates that the tag is at the beginning of a chunk that follows another chunk without O tags between the two chunks.
    \item the I-prefix indicates that the tag is inside a chunk. 
    \item the O-prefix indicates that the token belongs to no chunk.
\end{enumerate}

  In this step, a BiLSTM+CRF model and a CNN-BiLSTM model are deployed to achieve the goal and a comparison of performance is taken to demonstrate the difference, advantages, and disadvantages between them.

\subsection{Temporal Relations and Medication Candidates Classification}
 Relation extraction is the task of predicting attributes and relations for entities in a sentence. Extracted relations usually occur between two or more entities of a particular type (e.g. Medication, Dosage) and fall into several semantic categories. In this work, these relations that are discussed are between medications and corresponding dates. For instance, in the example text from Figure \ref{fig:example_text_EMRs}, there are medication ``Methotrexate'' that happened between ``May 2019'' and``February 2020''.
 Generally, there are eight types of temporal relations, including before, after, simultaneous, overlap, begun\_by, ended\_by, during, and before\_overlap suggested by \cite{sun20122012}. However, only the types of before, after, and overlap meet our task.
 A BERT-based model with CNN as an additional learning layer is implemented for this step.

\subsection{Post-processing}

 After the phrases of entity recognition and relation extraction, to make the result more robust and representative, this post-processing step is to convert results to a structured  table or CSV shape.
 However, in the previous 2012 i2b2 NLP challenge on temporal relation extraction, post-processing was considered a set of classification routines for the EVENTs only, and the TIMEx catergory remained unclassified \cite{Lin2013MedTime:Narratives}. 
 In this work, we apply an open resource tool SparkNLP\footnote{\url{https://github.com/JohnSnowLabs/spark-nlp}}, combined with our medication entity recognition and temporal relation extraction system to generate more informative results and a structured representation table e.g. the example Table \ref{tab:my_label_example_output_table} generated from the example text in Figure \ref{fig:example_text_EMRs}.

\begin{table}
    \centering
    \setlength{\tabcolsep}{0.6mm}{
    \small
        \caption{Extraction Result Example}
\begin{tabular}{|c|c|c|c|c|}
\hline
ID &Event & Statues & Start & Stop\\ \hline
134529565 & Methotrexate & ON & May 2019&February 2020\\ \hline
134529566 & Methotrexate & OFF & February 2020 & Unknown\\ \hline
\end{tabular}
}    \label{tab:my_label_example_output_table}
\end{table}

\section{Experimental Evaluations}
\label{chapt_eval}

\subsection{Dataset and Evaluation Metrics}

 In this work, the i2b2 challenge track 2009 and 2012 datasets were used. The objective is to gather details about every drug that the patient is known to take for each patient report that is given. 
 Partners Healthcare's discharge summaries will be the input for the medication challenge \cite{sun20122012}. These files are loosely organised. Many drugs are covered in the narrative text in addition to being listed by name in lists of pharmaceuticals. 
 According to statistical analysis, it contains 447 EMRs, with 252 annotated, which could be used to train and test the model. 
 While the data from the i2b2-2012 temporal relations challenge has similar content as the challenge track of 2009 does, it has a different format, additional entity categories, and temporal relations between events. 
 It consists of 760 files including 190 TXT files (texts), 190 XML.EXTENT files (annotation and position in original text), XML files (annotation and text) and TLINK files (relations) \cite{gurulingappa2012development}.
 The data sets are divided into 70\%, 15\% and 15\% for training, validation, and testing, respectively on both these two i2b2 challenges.
 The entity tags used in the i2b2 challenge track 2009 and 2012 have some differences which are shown in Table \ref{table:example_i2b2_2009} and Table \ref{table:example_i2b2_2012}.

\begin{table}
    \centering
    \caption{Tags Used in the i2b2 2009 Dataset.}
    \label{table:example_i2b2_2009}
    \begin{tabular}{ccc}
    \hline
       Tag &	Meaning & Example \\
      \hline
       m &  medication & Percocet \\
      \hline
      do &  dosage & 3.2mg \\
      \hline
      f &  frequency & twice a day \\
      \hline
      mo &  mode (/route of administration) & Mode: “nm” (not “Tablet”)\\
      \hline
      du &  duration & 10-day course\\
      \hline
      r &  reasons & Dizziness\\
      
      \hline
    \end{tabular}
  \end{table} 
  
  \begin{table*}
    \centering
    \caption{Tags Used in the i2b2 2012 Dataset.}
    \label{table:example_i2b2_2012}
    \begin{tabular}{ccc}
    \hline
       Tag &	Meaning & Example \\
      \hline
       CLINICAL\_DEPT &  clinical department & emergency room \\
      \hline
      EVIDENTIAL &   events that have an ‘evidential’ nature &  CT \textbf{shows} \\
      \hline
       TEST & clinical tests & CT \\
      \hline
      PROBLEM & symptoms  & sickness\\
      \hline
      TREATMENT &  medications, surgeries and other procedures & Levaquin\\
      \hline
      OCCURRENCE & the default value for other event types & He was \textbf{readmitted} for\\
      
      \hline
    \end{tabular}
  \end{table*}

 The evaluation metrics we used are Precision, Recall, F1 score, and Accuracy based on \textit{exact matching}. 
 Considering that there are multiple categories of labels, two different kinds of evaluation criteria are applied: 1) \textbf{macro average} evaluation, the function to compute precision, recall and F1-score, for each label and returns the average without considering the proportion for each label in the dataset; and 2) \textbf{weighted average} evaluation, the function to compute precision, recall and F1-score for each label, and returns the average considering the proportion for each label in the dataset.

\subsection{Model-I: BiLSTM-CRF for NER Task}


Combined with the DL model BiLSTM and graph model CRF, we carried out the NER task of electronic diseases using BiLSTM-CRF. The specific steps carried out are as follows:
\begin{enumerate}
\item Apply the Glove or bag-of-word model to obtain words' semantic representation.
\item 
Conduct sequence learning 
using the BiLSTM network, which effectively avoids the gradient disappearance and explosion of other DL models.
\item Restrict the sequence relationship between tags by using CRFs.
\end{enumerate}

\begin{figure*}
    \centering
    \includegraphics[width=1\textwidth]{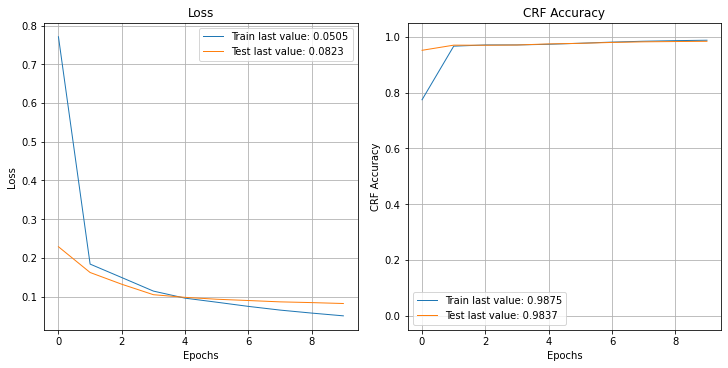}
    \caption{BiLSTM-CRF Model Training Process on i2b2-2009 Dataset}
    \label{fig:2009-bilstm-train-process}
\end{figure*}

\begin{figure*}[h]
    \centering
    \includegraphics[width=1\textwidth]{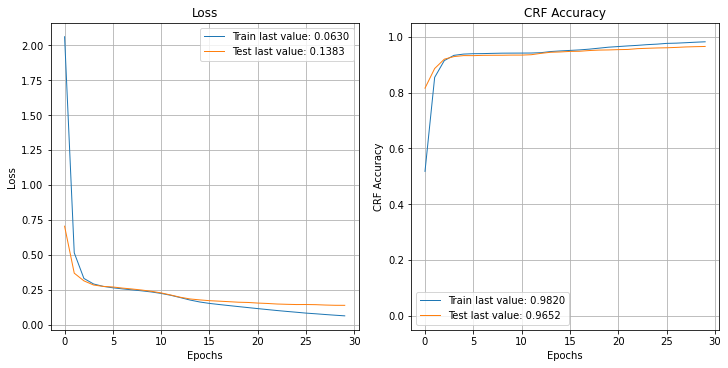}
    \caption{BiLSTM-CRF Model Training Process on i2b2-2012 Dataset}
    \label{fig:2012-bilstm-train-process}
\end{figure*}

    Using one-hot word embedding, the evaluation scores on 2009 and 2012 data sets 
    are shown in Table \ref{tab:BiLSTM-CRF-2009-eval} and \ref{tab:BiLSTM-CRF-2012-eval} with the explanations of each tag in Table \ref{table:example_i2b2_2009} and \ref{table:example_i2b2_2012}. 
    The model parameters for 2012 data is listed in Table \ref{tab:2012CRFPas}. The model learning parameters for 2009 data is the same except for Max Epochs which is set to be 10,  and the number of total trainable parameters is 1,292,188 instead of 506,138.
    The learning curves on Accuracy and Loss using 2009 and 2012 data are displayed in Figure \ref{fig:2009-bilstm-train-process} and \ref{fig:2012-bilstm-train-process}, which have similar patterns.

      \begin{table}[t]
      \centering
            \caption{Model Parameters of BiLSTM+CRF for the 2012 Dataset}
      \begin{tabular}{c|c}
            \hline
        parameter   & value \\
          \hline
          DENSE\_EMBEDDING & 50\\
          LSTM\_UNITS & 50 \\
          LSTM\_DROPOUT & 0.2 \\
          DENSE\_UNITS & 100 \\
          BATCH\_SIZE & 256 \\
          MAX\_EPOCHS & 30 \\
          LEARNING\_RATE & 0.0001\\ \hline 
          Total Param & 506,138 \\
          \hline
      \end{tabular}
      \label{tab:2012CRFPas}
  \end{table}


  \begin{table}[t]
    \centering
    \caption{NER Evaluation on the 2009 Dataset with BiLSTM-CRF.}
    \label{table:example_table}
    \begin{tabular}{ccccc}
      \hline
      \multirow{2}{*}{category} & \multicolumn{2}{r}{Number / \%} & \multicolumn{2}{r}{Number  } \\
                                   & precision & recall & f1-score & support
 \\
	  \hline
      PADDING  &  99.99 & 1.00 & 1.00 & 111422 \\
      B-do  & 81.88 & 70.52 & 75.78 & 519 \\
      I-do  & 84.09 & 74.66 & 79.10 & 446 \\
      B-m  & 81.35 & 66.51  & 73.19 & 1069 \\
      I-m  & 87.60 & 70.42 & 78.07 & 622 \\
      B-f & 83.50 & 77.48 & 80.37 & 444\\
      I-f & 75.44 & 61.43 & 51.81 & 166 \\
      B-du & 61.11 & 15.28 & 24.44 & 72\\
      I-du & 44.44 & 17.78 & 25.40 & 135\\
      B-r & 39.80 & 20.42 & 26.99 & 191\\
      I-r & 41.67 & 10.64 & 16.95 & 141\\
      B-mo & 83.05 & 78.61 & 80.77 & 374\\
      I-mo & 0.00 & 0.00 & 0.00 & 20\\
      o & 95.87 & 98.50 & 97.17 & 32429\\
      \hline
      accuracy & 98.63 & 148050 \\
      macro avg & 68.56 & 53.76 & 58.55 & 148050 \\
      weighted avg & 98.46 & 98.63 & 98.50 & 148050\\
      \hline
    \end{tabular}
    \label{tab:BiLSTM-CRF-2009-eval}
  \end{table} 
  
  \begin{table}[t]
    \centering
    \caption{NER Evaluation on the 2012 Dataset with BiLSTM-CRF.}
    \label{table:example_table10}
    \begin{tabular}{ccccc}
      \hline
      \multirow{2}{*}{category} & \multicolumn{2}{r}{Number / \%} & \multicolumn{2}{r}{Number  } \\
                                   & precision & recall & f1-score & support
 \\
	  \hline
      PADDING  &  1.00 & 99.99 & 99.99 & 63867 \\
      B-CLINICAL\_DEPT & 68.85 & 32.81 & 0.44 & 128\\
      I-CLINICAL\_DEPT & 80.43 & 54.95 & 65.29 & 202 \\
      B-EVIDENTIAL & 1.00 & 4.11 & 7.89 & 73 \\
      I-EVIDENTIAL & 0.00 & 0.00 & 0.00 & 6 \\
      B-OCCURRENCE & 61.69 & 44.39 & 51.63 & 410\\
      I-OCCURRENCE & 34.69 & 7.46 & 12.27 & 228 \\
      B-PROBLEM & 59.10  &  56.64 &  57.85  &     602\\
      I-PROBLEM  &   75.67  &  61.92  &  68.11   &    864\\
      B-TREATMENT  &   66.80  &  68.31  &  67.55    &   486\\
      I-TREATMENT  &  65.87 &   67.63  &  66.74   &    448\\
      B-TEST   &  58.48 &   54.18  &  56.25    &   299\\
      I-TEST  &   64.42  &  63.23    &63.82    &   378\\
      O   &  87.82  &  95.83 &    91.65 &  75.82\\
      \hline
      accuracy & 97.11  & 75573 \\
      macro avg & 65.99  &  50.82  &  53.82  & 75573 \\
      weighted avg & 96.89 &   97.11  &  96.88 & 75573\\
      \hline
    \end{tabular}
        \label{tab:BiLSTM-CRF-2012-eval}
  \end{table}

 As mentioned, evaluation metrics were based on exact-matching criteria, and with this setting, the BiLSTM+CRF model achieves a 98.50 weighted average F1 and 58.55 of Macro average F1 on 2009 data. 
 Apparently, combined with the evaluation scores of each subcategory, it can be seen that the amount of entities of each category varies significantly due to the uneven data distribution, which leads to the scores between entity categories varying wildly.

  As demonstrated in Table \ref{tab:BiLSTM-CRF-2009-eval}, the BiLSTM+CRF model has better prediction performances on class ``B-mo'', ``B-f'', ``I-do'', ``I-m'', ``B-do'' in addition to ``o'' with 0.75+ F1. and ``B-m'' in the 2009 dataset. The classes that have under 0.50 F1 scores also have much fewer supporting labels, less than 200.
  Similar patterns can be observed in the experiments on 2012 data set, i.e. higher number of supporting labels produce high F1 scores. 

   However, there are some \textit{kind of} exceptions, for instance, ``I-r'' and ``B-du'' have the supporting labels (141, 72) but their corresponding Macro Avg scores are (16.95, 24.44).
    In addition, the results based on precision, recall, and F1-score are very different. For most sub-classes, the model has higher precision than recall values which means it returns fewer results, but most of its predicted labels are correct when compared to the gold labels. 
    Namely, the percentage of Negative labels is lower.

 In order to explore the impacts of different embedding inputs for the BiLSTM+CRF model, we trained the model with Glove (glove.6B.50d), word2vector, in addition to one-hot encoding, and the average and standard deviations of metrics are recorded. 
 Table \ref{tab:BiLSTM-CRF-2009-diff-embeddings} demonstrates the performance of each model on the same dataset using Macro average F1 score. 
 Glove embedding achieved 72.48\% of precision, 52.58\% of recall, and 60.86\% of f1-score, respectively which increased by around 2\% than one-hot embedding. 
 For word2vec embedding, this model achieved exact 73.62\% precision, 60.34\% of recall, and 61.48\% of f1-score achieving a further improvement of 1.14\% in precision,  7.76\% in recall and 0.62\% than Glove embedding.

\begin{table}[]
    \centering
    \caption{Macro Avg Results of Different Embeddings on the 2009 Test Set using BiLSTM+CRF.}
    \begin{tabular}{cccc}
      \hline
    embedding & precision & recall & f1-score\\
    \hline
      one-hot & 68.56 & 53.76 & 58.55 \\
      Glove & 72.48  &  52.58  &  60.86   \\
      word2vec & 73.62 &   60.34  &  61.48 \\
      \hline
    \end{tabular}
    \label{tab:BiLSTM-CRF-2009-diff-embeddings}
  \end{table} 

\begin{table}
    \centering
    \caption{Macro Avg Results of Experiments on the 2009 Test Dataset with CNN-BiLSTM.}

    \begin{tabular}{cccc}
      \hline
    model & precision & recall & f1-score\\
    \hline
      CNN-BiLSTM & 75.67 & 77.83 & 78.17 \\
      \hline
    \end{tabular}
    \label{tab:CNN-BiLSTM-2009}
  \end{table} 

\subsection{Model-II: CNN-BiLSTM for NER}
 
The Convolutional Neural Network (CNN) used in this approach draws inspiration from \cite{chiu2016named} hybrid model, which uses bidirectional LSTMs to learn both character and word-level information. As a result, our model employs word embeddings, in addition to other human-created word features and character-level information retrieved using a convolutional neural network. All of these features are sent into a BiLSTM for each word token.
Following the tutorial by \cite{CNN-BILSTM}, we adapted the code from \cite{chiu2016named}, to implement the CNN-BiLSTM model for NER task on the 2009 dataset.

The idea of this model is to perform NER prediction by CONCAT of both character-level and word-level features and their combined input knowledge into BiLSTM learning. This enables the model to make better use of character-level features such as prefixes and suffixes, which can reduce the work of manual feature construction. 
The character-level features are obtained by a CNN model which has been approved as a great success in the NER task and POS tagging task \cite{chotirat2021part,labeau2015non}. 
Per-character feature vectors, such as character embeddings and character kinds, are concatenated with a max layer to produce a new feature vector, which is then used to train a model for recognising words.

The BiLSTM layer forms the core of the network and has the following three inputs: character-level patterns identified by CNN, word-level input from GloVe embeddings, and casing input (word case changes).
One of the critical tasks here is feature fusion as mentioned by \cite{chiu2016named}. 
Additional features can be applied 
for example, words can be divided into six categories, including all uppercase letters, all lowercase letters, all numbers, some numbers, and so on. 
The model includes the following 7 layers of network structure and their functions.
\begin{enumerate}
    \item Character embedding layer, which maps the input of characters to 30 dimensional embedding.
    \item Dropout layer (0.5), which mitigates the effect of overfitting
    \item1D convolutional layer, which transforms the character dimension size into 1
    \item Word embedding layer, which maps the words into 50 dimensional embedding vectors with Glove (glove.6B.50d)
    \item Concatenation layer, which combines processed character-level, word-level and casing data into a vector of 80 dimensions
    \item BiLSTM layer: using the merged word vector sequence as input, the spatial semantic modelling of the preceding and subsequent text information was carried out, and the bidirectional semantic dependence of word vectors was captured. The high-level feature expression of the context information of medical records was further constructed
    \item Dense output layer, which applies softmax function for prediction.
\end{enumerate}

The model parameters of CNN-BiLSTM is displayed in Table \ref{tab:CNNBILSTM-parameter} 
with codes adapted from `github.com/kamalkraj'\footnote{\url{https://github.com/kamalkraj/Named-Entity-Recognition-with\-Bidirectional-LSTM-CNNs}}.
This CNN-BiLSTM model was set for training for 30 Epoch but convergence after 14 Epoch. 
Table \ref{tab:CNN-BiLSTM-2009} presents the performance macro average precision, recall and f1-score of this model which is better than the scores from the BiLSTM-CRF model. The most increase comes from recall score from 52.58 to 77.83 using both Glove embedding, leading to the F1 score increase from 60.86 to 78.17.

\begin{table}[]
      \centering
      \caption{Model Parameters of CNN-BiLSTM for the i2b2 2009 Data}
      \begin{tabular}{c|c}
            \hline
        parameter   & value \\
          \hline
          DENSE\_EMBEDDING & 30\\
          DROPOUT & 0.5 \\
          WORDS\_EMBEDDING & 50 \\
          DROPOUT\_RECURRENT & 0.25 \\
          LSTM\_STATE\_SIZE & 200 \\
          CONV\_SIZE & 3   \\
          LEARNING\_RATE & 0.0105\\
          OPTIMIZER & Nadam()  \\
          \hline
      \end{tabular}
      \label{tab:CNNBILSTM-parameter}
  \end{table}

\begin{table}[t]
      \centering
      \caption{Model Parameters of BERT+CNN }
      \begin{tabular}{c|c}
            \hline
        parameter   & value \\
          \hline
          BERT\_MODEL & $Bio-BERT_{base}$\\
          DROPOUT & 0.5 \\
          EPOCH & 5 \\
          DROPOUT & 0.1 \\
          MAX\_SENTENCE\_LENGTH & 512 \\
          HIDDEN\_SIZE & 768   \\
          LEARNING\_RATE & 1e-05\\
          \hline
      \end{tabular}
      \label{tab:BERTCNN-parameter}
  \end{table}

\subsection{Model-III: BERT-CNN for Temporal Relation Extraction}

\begin{figure}[t]
    \centering
\includegraphics[width=0.5\textwidth]{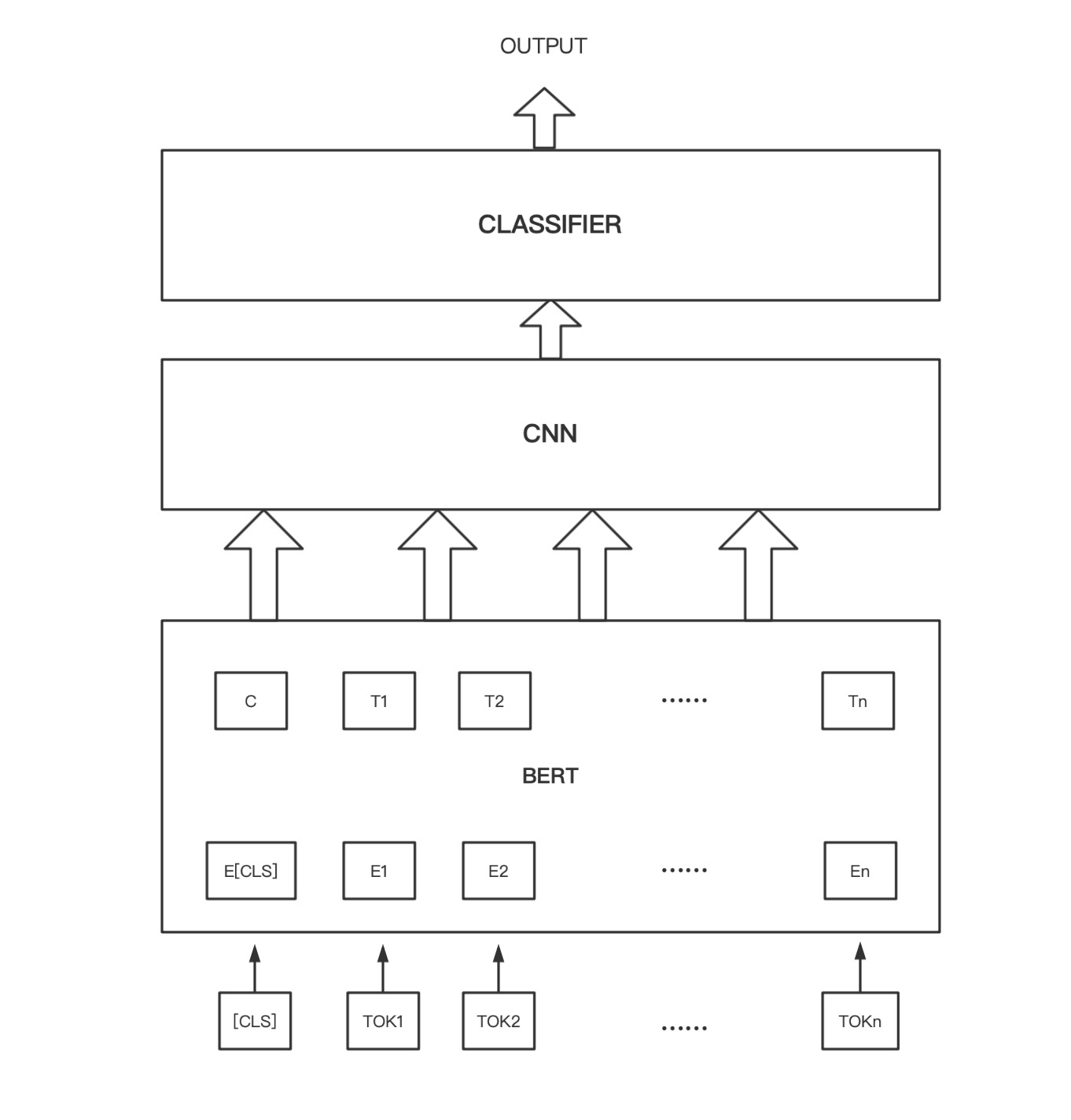}
    \caption{The Architecture of BERT-CNN Model  Re-Implemented}
    \label{fig:BERT-CNN-structure} 
\end{figure}

 Although the [CLS] tags in BERT output can achieve reasonable classification results, the rich semantic knowledge contained in BERT is not fully utilised. 
 Therefore, in this work, inspired by \cite{michalopoulos2020umlsbert}, the CNN model is fused to expand BERT learning.
 Code adapted from open source projects of UmlsBERT\footnote{\url{https://github.com/gmichalo/UmlsBERT}} and ``MedicalRelationExtraction''\footnote{https://github.com/chentao1999/MedicalRelationExtraction}, we trained this BERT+CNN model based on the parameters on the i2b2 challenge 2012 dataset. 
The parameters in BERT-CNN model are set as in Table \ref{tab:BERTCNN-parameter} and the model structure is presented in Figure \ref{fig:BERT-CNN-structure}.
 After the BERT model receives the processed text, the content of the EMRs text is represented by a vector through the two-layer Transformer mechanism of the BERT model. 
 The model produces a vectorised representation of comprehensive semantic information fused by each token vector, sentence vector, and feature vector in the EMRs and then feeds the output of the model to the CNN layer. 
 In this work, three different convolutional kernels are used to capture different feature information. After the fully connected layer is connected by word vector mapping, the CNN model further extracts the semantic information of the dialogue text.


\textbf{Data Sampling and  Evaluation:}
 In order to mitigate the influence of the training dataset imbalance of each sub-class, we down-sampled the dataset to 3000 samples for each class (AFTER, OVERLAP and BEFORE). Table \ref{table:BERT-CNN-Temporal-2012-train-sample} gives the result of the prediction on the training dataset. Table \ref{table:BERT-CNN-Temporal-2012-test} shows the result of $BERT_{base}+CNN$ model performance for the relation extraction (RE) task on the i2b2-2012 test data. 
 This model achieves a macro avg. 64.48\% of precision, 67.17\% of recall, and 65.03\% of F1-score.

\begin{table}[]
    \centering
    \caption{Temporal RE Evaluation on the 2012 Sampled Training Dataset using BERT-CNN.}
    \label{table:BERT-CNN-Temporal-2012-train-sample}
    \begin{tabular}{ccccc}
      \hline
      \multirow{2}{*}{category} & \multicolumn{2}{r}{Number / \%} & \multicolumn{2}{r}{Number  } \\
                                   & precision & recall & f1-score & support
 \\
	  \hline
      AFTER  &  96.23 & 97.74 & 96.86 & 3000 \\
      OVERLAP & 93.71 & 95.85 & 94.80 & 3000\\
      BEFORE & 97.13 & 93.24 & 95.06 & 3000 \\
      \hline
      accuracy & && 95.60  & 9000 \\
      macro avg & 95.60  &  95.60  &  95.60  & 9000 \\
      weighted avg & 95.60 &   95.60  &  95.60 & 9000\\
      \hline
    \end{tabular}
  \end{table} 

\begin{table}[t]
    \centering
    \caption{Temporal RE Evaluation on the 2012 Test Dataset using BERT-CNN}
    \label{table:BERT-CNN-Temporal-2012-test}
    \begin{tabular}{ccccc}
      \hline
      \multirow{2}{*}{category} & \multicolumn{2}{r}{Number / \%} & \multicolumn{2}{r}{Number  } \\
                                   & precision & recall & f1-score & support
 \\
	  \hline
      AFTER  &  34.52 & 49.00 & 40.52 & 1122 \\
      OVERLAP & 67.34 & 77.23 & 71.94 & 4078\\
      BEFORE & 91.62 & 75.28 & 82.64 & 6000 \\
      \hline
      accuracy & &&73.32  & 11200 \\
      macro avg & 64.48  &  67.17  &  65.03  & 11200 \\
      weighted avg & 77.03 &   73.34  &  74.47 & 11200\\
      \hline
    \end{tabular}
  \end{table}

 As the prediction shows, with BERT-base as a word vector extraction or directly as a classification model, the BERT+CNN model can achieve a relatively good classification result, indicating that the pre-trained model BERT-base can extract the semantic information well of dialogue text. 
 Although the BERT+CNN model has reasonable performance for relation extraction, over-fitting on the training dataset is still inevitable, especially for sub-class AFTER and OVERLAP, which show 61.71\% and 26.37\% lower than the prediction results on the training dataset, respectively. 
 There is still the correlation between the number of test data samples and prediction precision which is that the more supporting samples, the better the prediction performance, which is similar to our NER findings.

 \subsection{Post-Processing}
 \label{appendix_sub_Model_Analysis/Post-pro}
 In order to apply the prediction results of the NER task for medication recognition and the RE task for temporal relation extraction between medication event and date on medication usage status, we design a set of rules that determine whether a medication is in use. 
 We take the date of Admission and Discharge as a time-frame and take the relation between the time-frame and medication into consideration. The dataset texts come from de-identified discharge summaries (HOSPITAL COURSE), which give the treatment information of patients during hospital time. The temporal relation on medication is categorised into several situations for `in use' and `not' as below.
\begin{enumerate}
    \item If the relation between Admission Date and a medication AFTER and the relation between Discharge Date and the medication is BEFORE or OVERLAP, it means the medication is in use.
    \item If the relation between Admission Date and a medication OVERLAP and the relation between Discharge Date and the medication is OVERLAP, it means the medication is in use.
    \item If the relation between a date except for Admission and Discharge Date which is between them and a medication is OVERLAP and the relation between Discharge Date and the medication is BEFORE or OVERLAP, it means the medication is in use.
    \item  Otherwise it is not.
    
\end{enumerate}

With this set of rules, we manually checked the classification results, showing that nearly three out of ten can be classified correctly. 

\section{Related Work}
NER and RE belong to the information extraction (IE) task, which has been developed using different methods, from rule-based ones, to machine learning and deep learning models.
The rule-based methods require expertise in designing hand-crafted rules suitable for specific tasks and domains. While these methods worked out in some situations, it is often difficult for other researchers to adopt the existing rules into a new experiment \cite{ji2019hybrid}.
Remarkable work in this category includes 
\cite{Reeves2012DetectingNarratives,Strotgen2013MultilingualTagging,Hao2018ATexts}
on temporal expression extractions, 
and \cite{Roberts2013AText,Lin2013MedTime:Narratives} on medical terminology mining.

Earlier stage ML models on TLINKS detection and clinical events and concepts extraction include \cite{Chang2013TEMPTINGSummaries,Xu2013AnChallenge,wu2015named,wu2017clinical} where CNNs and RNNs were explored.
The combined model of BiLSTM-CRF was also investigated by some existing work reported in \cite{kim2020korean,lee2020biobert,jiang2019bert} and in language-specific experiments such as Chinese EMRs by \cite{ji2019hybrid}.
Similar structures including BiLSTM-CNN-CRF and BiLSTM-CRF were investigated in general domain NER and POS tagging by \cite{ma-hovy-2016-end-bilstm-cnn-crf,lample-etal-2016-neuralNER}. 


Based on these previous findings, our work takes one step further and investigated various combinations of these state-of-the-art neural learning models on clinical domain NER and Temporal modelling.

\section{Conclusions and Future Work} 
\label{chapt_conclusion}

We carried out empirical study on medication extraction and temporal relation classification using three hybrid machine learning models BiLSTM-CRF, CNN-BiLSTM, and BERT-CNN. By combining these two tasks and a set of rules we designed, we aim to generate a simple structured output indicating if certain medications are in use during the extracted timeframe. 
While the experimental work demonstrated promising results, we plan to further improve the model performances from different aspects, which include the integration of domain-specific pertained BERT model such as Med-BERT \cite{rasmy2021med}, more text features, data-set augmentation especially for low frequency labels e.g. using \textit{synthetic data }\cite{Belkadi-etal-2023-lt3}, and system optimisation using different learning structures. We also investigate the \textit{prompt-based engineering} based on LLMs for MedTem2.0 project \cite{cui2023medtem2}.

\section*{Limitations}
  The main sources of our experimental data are the i2b2 Challenge 2009 and 2012, which are considerable corpora leading to a lack of computation resources to do more experiments for fine-tuning the best model settings and parameters. 
  Moreover, the proportion of data (those annotated) on the relationship between medications and their corresponding dates is inadequate, which can not provide sufficient information for MEDICATION-DATE relation extraction. 
  Therefore, the performance of information extraction generated by the rules created for patients' medication usage status falls short of ideal standards, which requires further improvement. 

\section*{Ethical Statement}
There are no ethical concerns in this work since the data sets we used from n2c2 challenges are anonymised and ethically approved by the shared task organisers.

\section*{Acknowledgements}
LH and GN are grateful to grant support EP/V047949/1 ``Integrating hospital outpatient letters into the healthcare data space'' (funder:
UKRI/EPSRC).  

\bibliography{nodalida2023}

\bibliographystyle{IEEEtran}

\end{document}